\documentclass{llncs}
\usepackage{makeidx}  % allows for indexgeneration

\usepackage{booktabs} % For formal tables}

\usepackage{glossaries-extra}
\usepackage{hyperref}
\usepackage{graphicx}
\usepackage{subcaption}
\usepackage{wrapfig}
\usepackage{float}
\usepackage{multirow}

\newabbreviation{aoi}{AOI}{Area-of-Interest}
\newabbreviation{fov}{FoV}{field of view}
\newabbreviation{gpgpu}{GPGPU}{General Purpose Graphics Processing Unit}
\newabbreviation{uoi}{UoI}{union over intersection}
\newabbreviation{acf}{ACF}{aggregated channel features}
\newabbreviation{dpm}{DPM}{deformable part model}
\newabbreviation{paf}{PAF}{part affinity fields}

\begin{document}
\frontmatter          % for the preliminaries
\pagestyle{headings}
\addtocmark{Hamiltonian Mechanics}

\title{Automated analysis of eye-tracker-based human-human interaction studies}

\author{Timothy Callemein\textsuperscript{1}, Kristof Van Beeck\textsuperscript{1}, Geert Br\^{o}ne\textsuperscript{2}, Toon Goedem\'{e}\textsuperscript{1}}

\institute{\textsuperscript{1}KU Leuven - EAVISE, Jan Pieter de Nayerlaan 5, Sint-Katelijne-Waver, BELGIUM\\
\textsuperscript{2}KU Leuven - MIDI, Sint-Andriesstraat 2, Antwerpen, BELGIUM}
\maketitle

\section*{Abstract}
Mobile eye-tracking systems have been available for about a decade now and are becoming increasingly popular in different fields of application, including marketing, sociology, usability studies and linguistics.
While the user-friendliness and ergonomics of the hardware are developing at a rapid pace, the software for the analysis of mobile eye-tracking data in some points still lacks robustness and functionality.
With this paper, we investigate which state-of-the-art computer vision algorithms may be used to automate the post-analysis of mobile eye-tracking data.
For the case study in this paper, we focus on mobile eye-tracker recordings made during human-human face-to-face interactions.
We compared two recent publicly available frameworks (YOLOv2 and OpenPose) to relate the gaze location generated by the eye-tracker to the head and hands visible in the scene camera data.
In this paper we will show that the use of this single-pipeline framework provides robust results, which are both more accurate and faster than previous work in the field.
Moreover, our approach does not rely on manual interventions during this process.

\keywords{Post analysis, Human-Human-interaction studies, mobile eye-trackers}

\section{Introduction}
\label{sec:introduction}

A growing field of application for mobile eye-trackers is the recording of human-human interactions, enabling researchers in the fields of linguistics and conversation analysis to analyse the role of eye gaze in non-verbal communication and interaction management.
This research, among others, focuses on the distribution of gaze of each interlocutor during face-to-face interactions, answering basic research questions such as: ``How long does a person spend looking at the face or hands of an interlocutor during a conversation?'', ``Does the distribution of visual attention differ depending on the type of interaction, the role or status of the participants, or other factors?''
Mobile eye-trackers contain the necessary hardware to simultaneously record the scene from the wearer's perspective and track the gaze of the wearer during this recording, providing an insider's perspective on the interaction.

Most software currently provided with mobile eye-trackers comes with a basic \gls{aoi} analysis method that allows the user to select a bounding box as \gls{aoi}, for example for determining specific objects of interest for an analysis of visual attention.
Afterwards the software matches the chosen \gls{aoi} to the current gaze location using the matching technique discussed in \cite{brone2011towards}.
This technique provides an automatic annotation of rigid objects containing similar features as the model.
During human-human interactions, however, the main \gls{aoi} would be faces of co-participants or hands performing gestures, but unfortunately both are non-rigid and impossible to recognize based on simple appearance-based techniques, such as in \cite{brone2011towards}.
The lack of an automatic annotation tool will leave most research in this field resorting to manual annotation of these types of 'objects'.
Depending on the amount of people taking part in the study, the amount of data grows, resulting in a cumbersome time-consuming annotation task.

\begin{figure}[bt]
 \centering
 \vspace{-5mm}
 \includegraphics[width=0.50\textwidth]{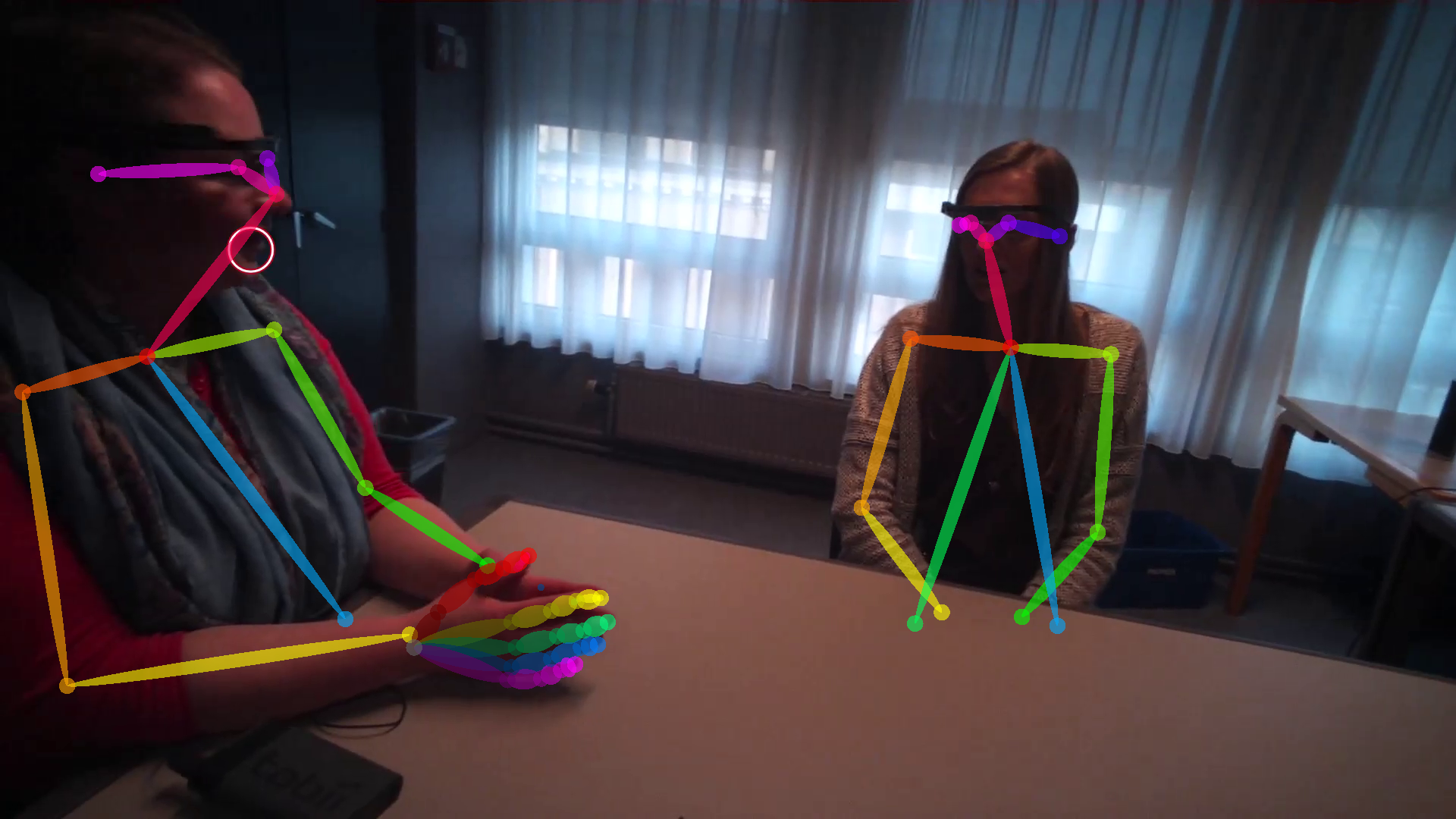}
 \caption{Frame from a three-persons-conversation processed by the pose estimator (coloured skeletons) with the gaze (red circle)\vspace{-5mm}}
 \label{fig:trivid_frame}
\end{figure}

In this paper we investigate if state-of-the-art computer vision techniques can perform accurate detection of hands and heads as the most relevant non-rigid objects for human interaction analysis, without making use of artificial markers.
These detections will, in a second step, be combined with the gaze coordinates of the eye-tracking camera, producing a fully automatic annotation tool which will eliminate a significant part of the manual annotation work. As an illustration, Figure \ref{fig:trivid_frame} shows a frame from a three-person-conversation (from the perspective of one of the co-participants wearing eye-tracking glasses) with (a) the output of a pose estimation algorithm \cite{cao2017realtime} and (b) a red circle representing the wearer's current gaze fixation.

The remainder of this paper is organized as follows.
In section \ref{sec:related_work} we discuss related work on previous post-analysis techniques with mobile eye-trackers, followed by three sections explaining our approach on detecting the torso in section \ref{sec:approach_torso}.
Based on the results of this initial detection step, we will advance to section \ref{sec:approach_head}, limiting the detection area to the head.
Section \ref{sec:approach_hands} focuses on the hands using the same techniques as explained in the previous sections.
In section \ref{sec:results} we discuss the results comparing previous work with both the YOLOv2 detector and pose estimation based detections. The paper closes off with a general conclusion and suggestions for future work in section \ref{sec:conclusion}.

\section{Related Work}
\label{sec:related_work}

Most eye-tracking systems are shipped with manufacturer software assisting the user in automated annotation and data aggregation, next to providing a manual annotation tool for more fine-grained analysis.
Mobile eye-trackers, while providing an advantage in mobility, come with the disadvantage of a dynamic spontaneous scene.
The most advanced manufacturerers ship software based on appearance-based image features (SIFT \cite{lowe2004distinctive}, SURF \cite{bay2006surf}) providing a rudimentary single-image-model annotation applicable for studies focussing on  rigid objects , but not be able to deal with non-rigid three-dimensional objects like hands and heads that change during the course of time and vary from person to person.

The work of De Beugher et al. \cite{de2014automatic} tries to answer the need for an automated annotation tool capable of annotating non-rigid objects by using a machine learned model on multiple images.
They use two techniques, one to train a more generalised model of a person's upper-body based on a \gls{dpm} \cite{felzenszwalb2010cascade}, while the second is trained to detect faces based on HAAR-features \cite{viola2001rapid}.
Both of these techniques are evaluated separately and combined in \cite{de2014automatic} on an annotated dataset.
For the purpose of this paper we will focus on the performance of the detector, leading to better results during gaze classification.
Furthermore, the upper-body comprises not only of the head location, offering no guarantee that the gaze placed upon the upper-body can be classified as the head due to its larger \gls{aoi}.

Apart from the head we want to automate the annotation process for hands that appear in the scene camera images generated by the eye-tracker, providing a first basic coding layer for e.g. gesture studies.
A hand detector combining a two-stage hypothesis and classification method by \cite{mittal2011hand} shows that a single model can be improved by taking other properties of our hands into account.
De Beugher et al. improved this work further in \cite{de2015semi} by adding their previous upper-body model as a preprocessing step.
Additionally, they implemented a rotating hand model and allowed manual interventions to further improve accuracy.
Yet, this model still suffers from the different orientations of the hand opposed to the orientation of the trained model.
Furthermore, they only achieve good results when sufficient manual interventions are given (1.61\% gives an F1-score above 85.17\%).
In our research we develop fully-automatic detectors with no obtrusive elements and no need for manual interventions.

Previously mentioned techniques involving machine learning seem insufficient to yield high accuracy in detecting a difficult object like the human hand.
These techniques are recently being overtaken by state-of-the-art neural network object detectors capable of extracting high levels of features that comprise an object.
Although deep learning solutions have been kept in the background for quite a while, recent advancements in \gls{gpgpu} hardware and an increase in available training data \textit{(e.g. ImageNet \cite{deng2009imagenet})} have allowed their emergence.
The current deep learning techniques greatly outperform previous hand-crafted and simple machine learning techniques.
In this paper we compare two state-of-the-art deep learning techniques and test their accuracy for the automatic annotation of mobile eye-tracking data.

\section{Torso detection}
\label{sec:approach_torso}

One of the baselines of human face-to-face interaction is the simple fact that interlocutors tend to gaze at the face of the other while interacting (with addressees typically gazing at the current speaker more and longer than the other way around \cite{oertel2012gaze,brone2017eye}).
This makes the head one of the prime objects to be detected as part of an automated annotation procedure.

Previous techniques tried to detect the head location by using the upper body or torso including the head \cite{de2014automatic}.
This allows the detector to use more information, leading to a better result.
In this paper we compare two state-of-the-art deep learning based techniques with traditional upper-body detectors.
In our research we have focussed on using the state-of-the-art YOLOv2 detector \cite{redmon2016yolo9000} based on the Darknet framework for retraining purposes.
We first annotated the torsos of 4000 images from the dataset provided by \cite{mittal2011hand}.
We then used this data by including pre-trained weights on the VOC-dataset \cite{pascal-voc-2009} to calculate new weights to detect the torso.

The second technique that we included, which is also based on deep learning,  is called pose estimation.
Pose estimators, compared to a conventional detector, will not only produce a bounding box around the person or detection in general.
They try to estimate the separate key-points of body-part-joints that together compose the pose of that person.
Using the key-points that are part of the torso, we can use the pose estimator as a torso detector, by returning a bounding box around them.
In this paper we have implemented the OpenPose framework \cite{wei2016cpm} bundling three components.
The first part is capable of detecting the separate anatomic body joint points (e.g. shoulder, elbows, wrists, ...).
When only one person is visible all the found points will belong to that person.
When there are multiple people in the image,  however, the body joint points will have to be grouped according to the person they belong to.
The second part includes a network capable of detecting the \gls{paf} between joints \cite{cao2017realtime}.
These \gls{paf} will assist the previous network in combining the joint points to the corresponding person.
The last part consists of detecting a more detailed pose of the hands, which will be discussed further in section \ref{sec:approach_hands}.

\section{Head detection}
\label{sec:approach_head}

\begin{figure}
\centering
\begin{subfigure}{.4\textwidth}
  \centering
  \includegraphics[width=0.95\linewidth]{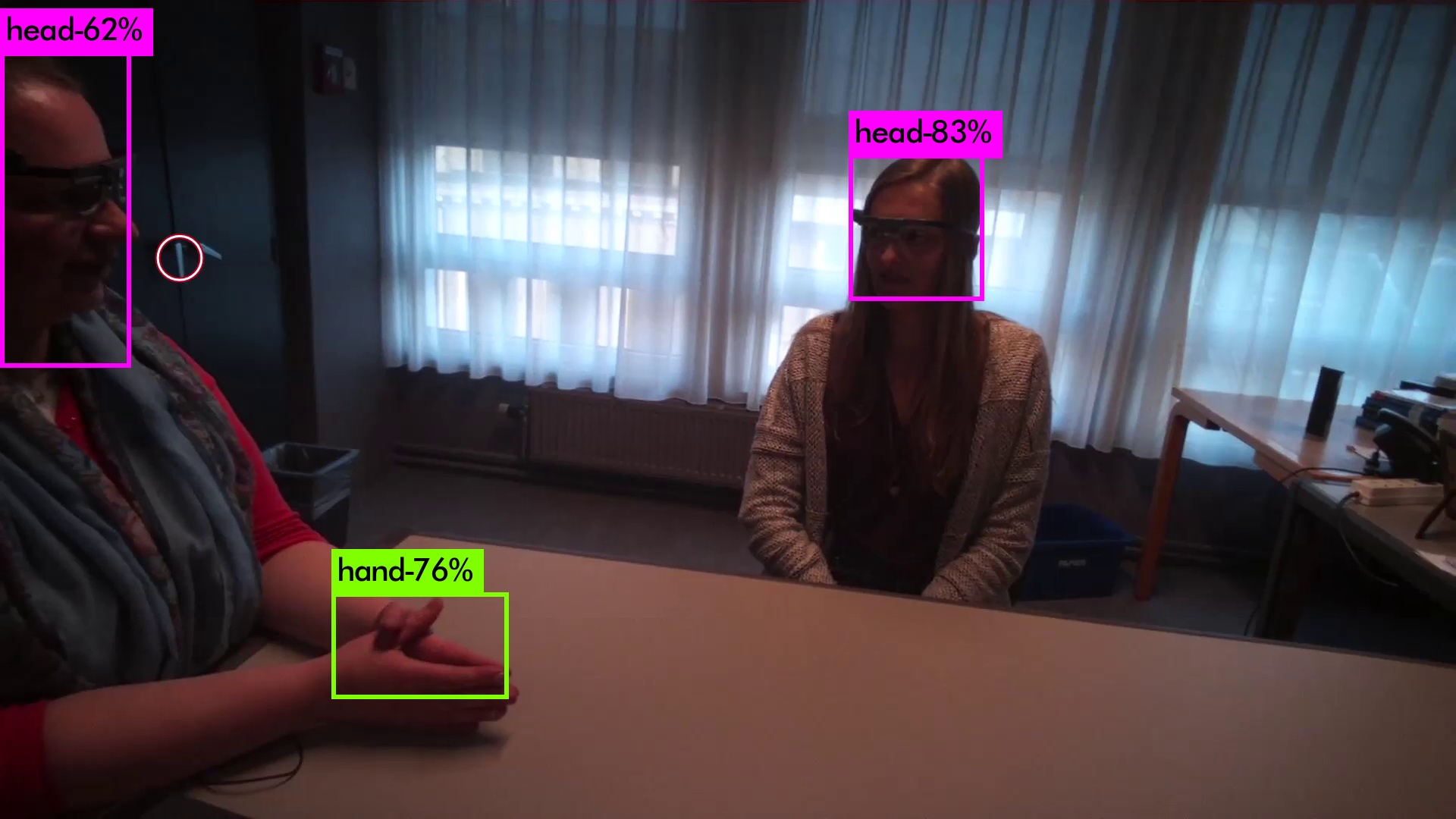}
  \caption{YOLOv2 based head/hand detection}
  \label{fig:head_yolo}
\end{subfigure}
\begin{subfigure}{.4\textwidth}
  \centering
  \includegraphics[width=0.95\linewidth]{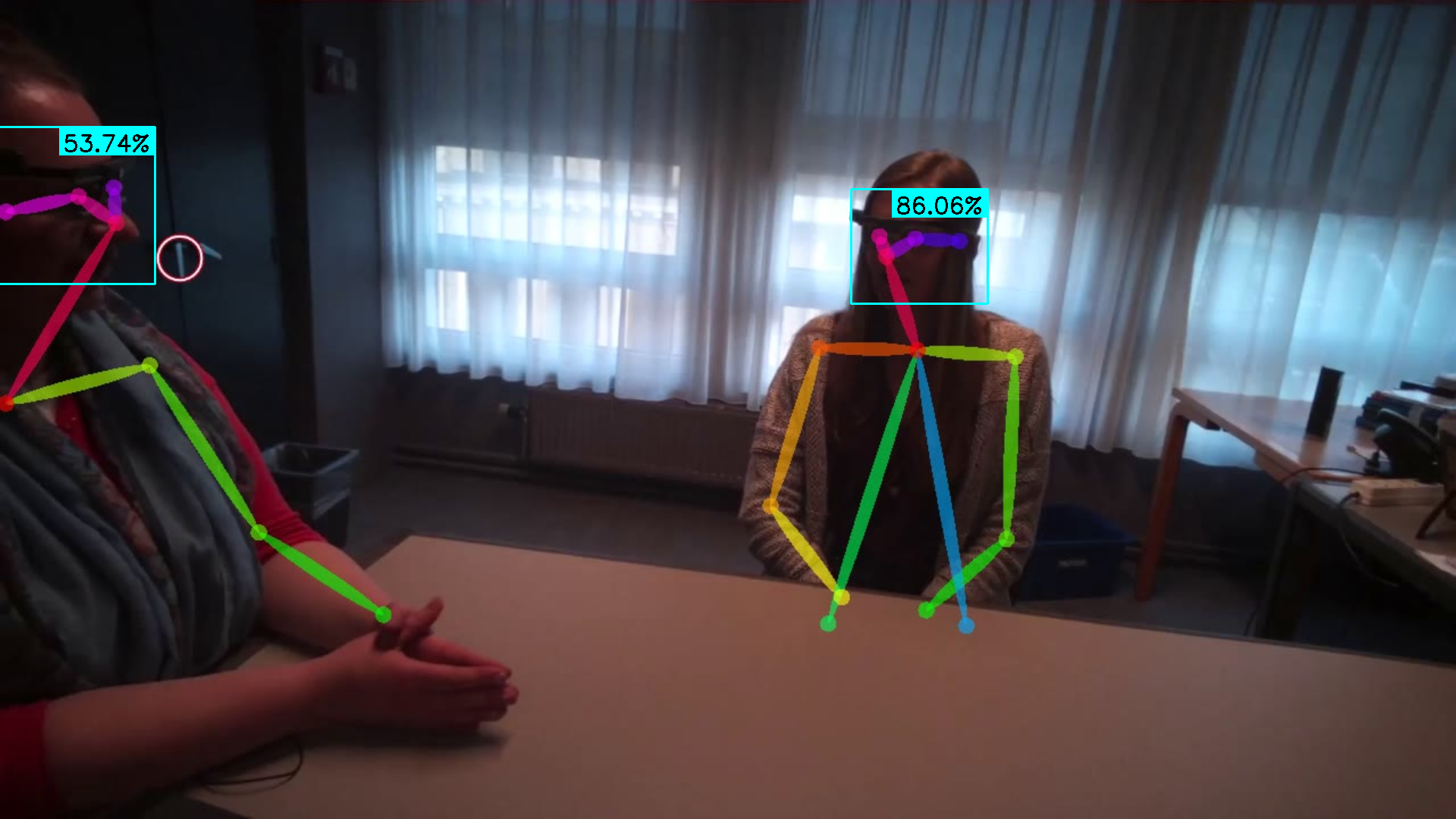}
  \caption{OpenPose based head detection}
  \label{fig:head_pose}
\end{subfigure}
\caption{Both techniques detecting the head}
\label{fig:head_plots}
\vspace{-8mm}
\end{figure}

In the previous section we used two state-of-the-art techniques in order to detect the torso.
As in \cite{de2014automatic}, this can be used to acquire context information on where to find the head.
Moreover, if one is only interested in the question whether the test person looks at another person, a torso detector suffices.
However, as mentioned above, researchers are generally interested in the question if the gaze is directed towards the face or head of the interlocutor.

In section \ref{sec:approach_torso}, we described how we retrained the YOLOv2 detector on a specific dataset in order to train a torso detector.
The retraining was based on around 1800 annotations from the manually annotated dataset used in \cite{simon2017hand}, containing both the head location and the hand key-points.
We have followed the same approach as with the torso-model to train a YOLOv2 model on this data, but included both the head and hand into a single detector.

Using the pose key-points we were able to find the torso.
However, these points are also usable to estimate the head position.
Yet the pose estimator only provides the eyes, ears and nose point.
We therefore determined a bounding box around the head based on these points by first looking at the head direction with respect to the camera.
When the head is frontal we return a bounding box based on the ear-by-ear distance and nose point.
However, when the head is in profile this is not possible.
Not only the centre will shift away from the nose, but some points will be self-occluded by the head.
By detecting the orientation of the head beforehand (e.g. frontal, right profile, left profile), we can provide a more accurate bounding box around the head.

Due to the margin of error presented by current mobile eye-trackers it is possible that the gaze cursor is focussed on the head, yet is not within the strict boundaries of the head.
This error margin may increase over time, especially with longer recordings with a single calibration step before the start.
During manual annotation, the final decision will depend on experience with the eye-tracker and general offset present on the eye-tracker.
Figure \ref{fig:head_plots} illustrates both the pose-based and YOLOv2-model based detections of the head with the gaze near the head, but not within the boundaries.
In our model we have included the option to increase the margins.
This allows for a bigger head boundary and a consistent annotation decision process.
However, in this paper during the evaluation of our detections in section \ref{sec:results} we used stricter boundaries to not influence our detection results.

\section{Hand pose estimation}
\label{sec:approach_hands}
Apart from the head, the hands also play a central role in non-verbal communication as prime articulators of visible bodily action.
We therefore compare different state-of-the-art detectors that may be useful as part of the annotation of eye-tracking data.
A particular challenge here is that fast motion of both the hands to be detected, as well as movements of the head by the person wearing the eye-tracking glasses may result in the hands being blurred and unclear in the images to be processed.
This contributes to the difficulty level of detecting them in an accurate way.
In section \ref{sec:approach_head} we trained a combined YOLOv2 model, hands and head, on a limited dataset.
Because of these challenges we decided to train a second dense hands-only model on around 18000 hand annotations taken from the \cite{mittal2011hand} dataset.
In the remainder of this paper, this detector is referred to as "YOLOv2 Dense".

Another hand pose estimator was presented by \cite{simon2017hand}.
They use the wrist location from the complete pose as a basis during the hand-pose estimation.
This model is capable of estimating each separate hand joint separately.

In order for the hand pose estimator to work, enough detail of the hand must be visible.
When the image is unclear or blurred we notice that the estimator fails.
We therefore developed an additional pose-based hand detector by using pose points of the arm.
A bounding box is estimated around the hand based on the length and direction of the vector between the wrist and elbow.

\begin{figure}[tb]
\centering
\begin{subfigure}{.32\textwidth}
  \centering
  \includegraphics[width=0.95\linewidth]{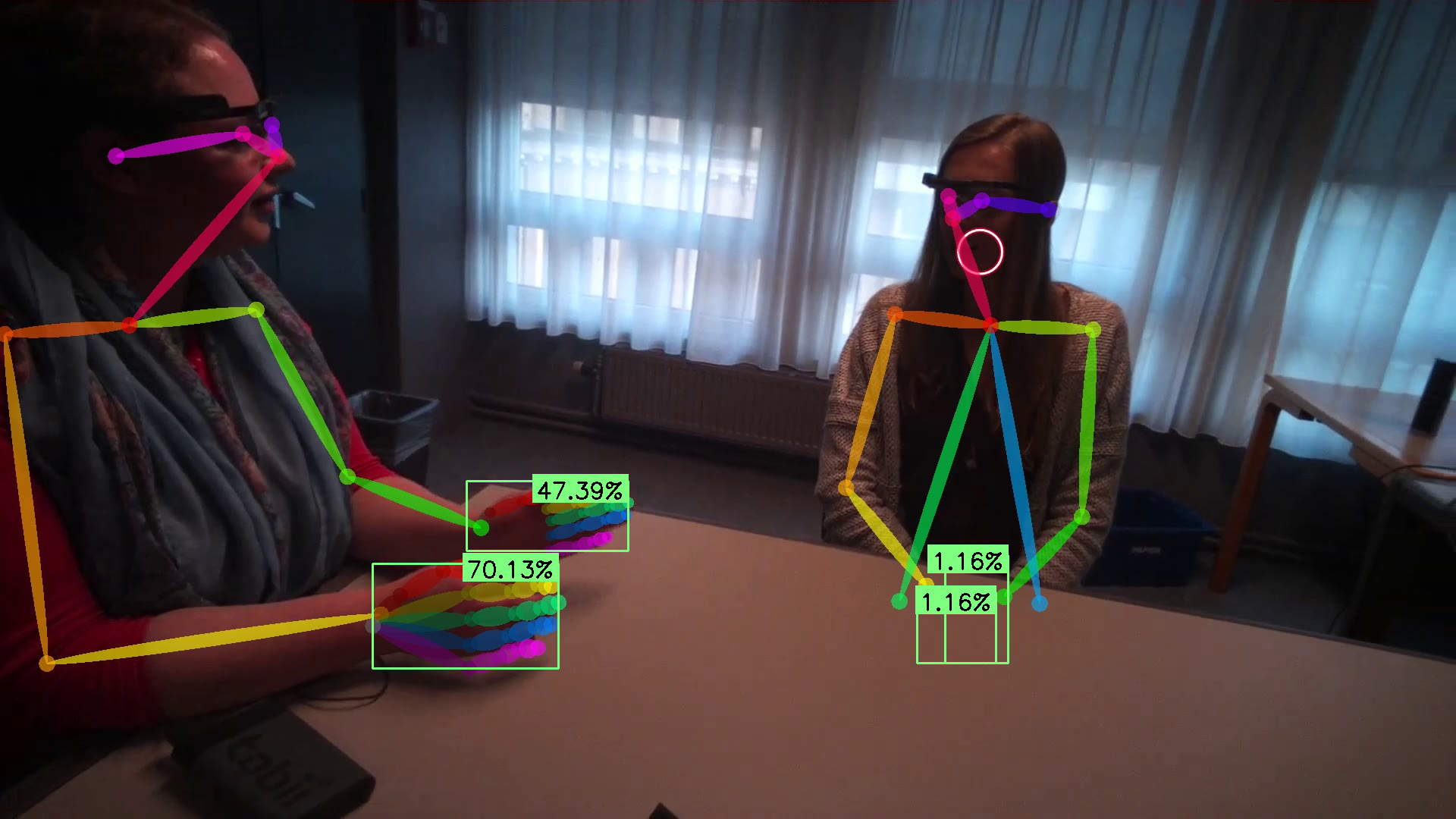}
  \caption{Hand pose based detection}
  \label{fig:hand_pose}
\end{subfigure}
\begin{subfigure}{.32\textwidth}
  \centering
  \includegraphics[width=0.95\linewidth]{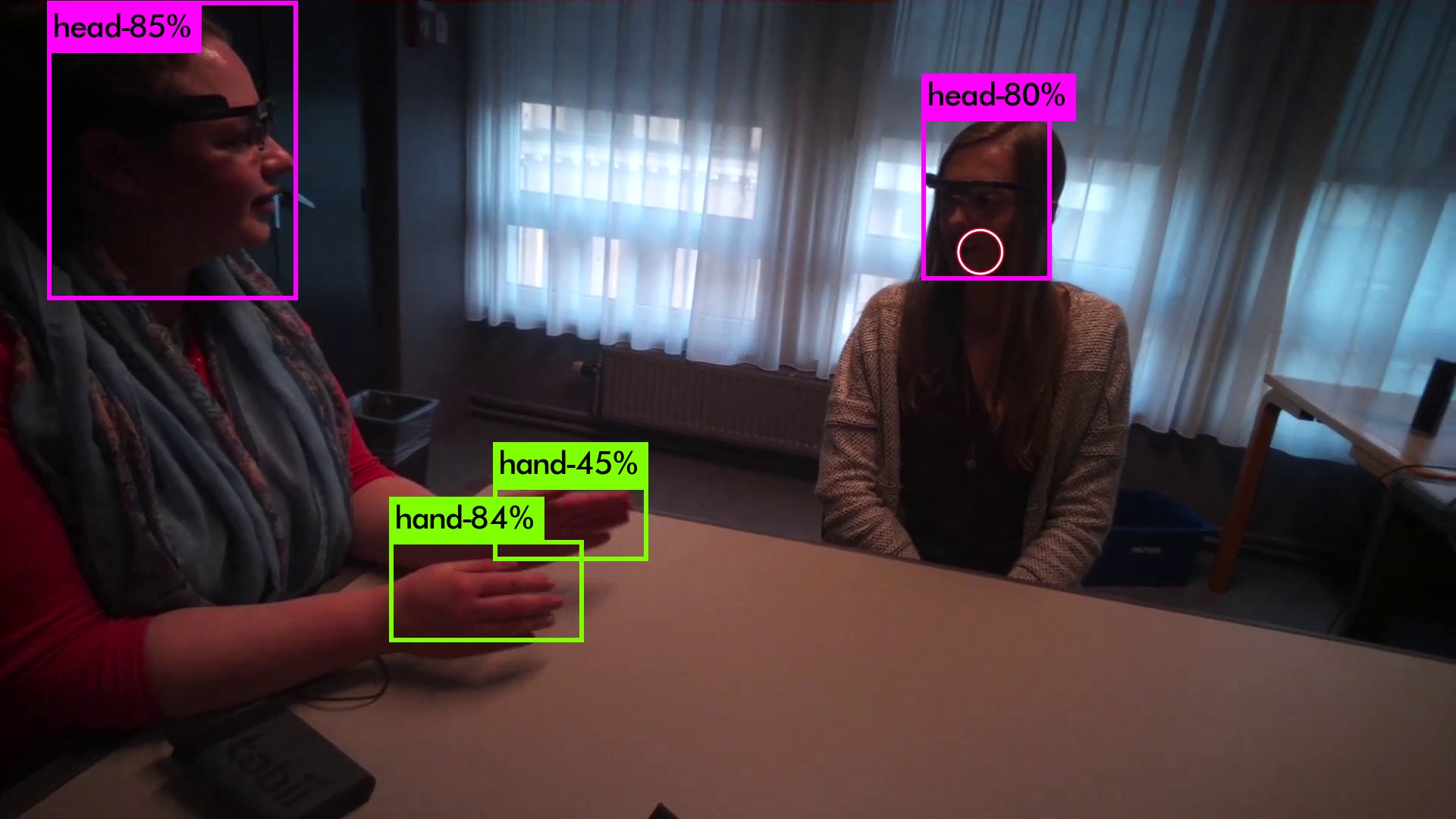}
  \caption{YOLOv2 Hand detection}
  \label{fig:hand_yolo}
\end{subfigure}
\begin{subfigure}{.32\textwidth}
  \centering
  \includegraphics[width=0.95\linewidth]{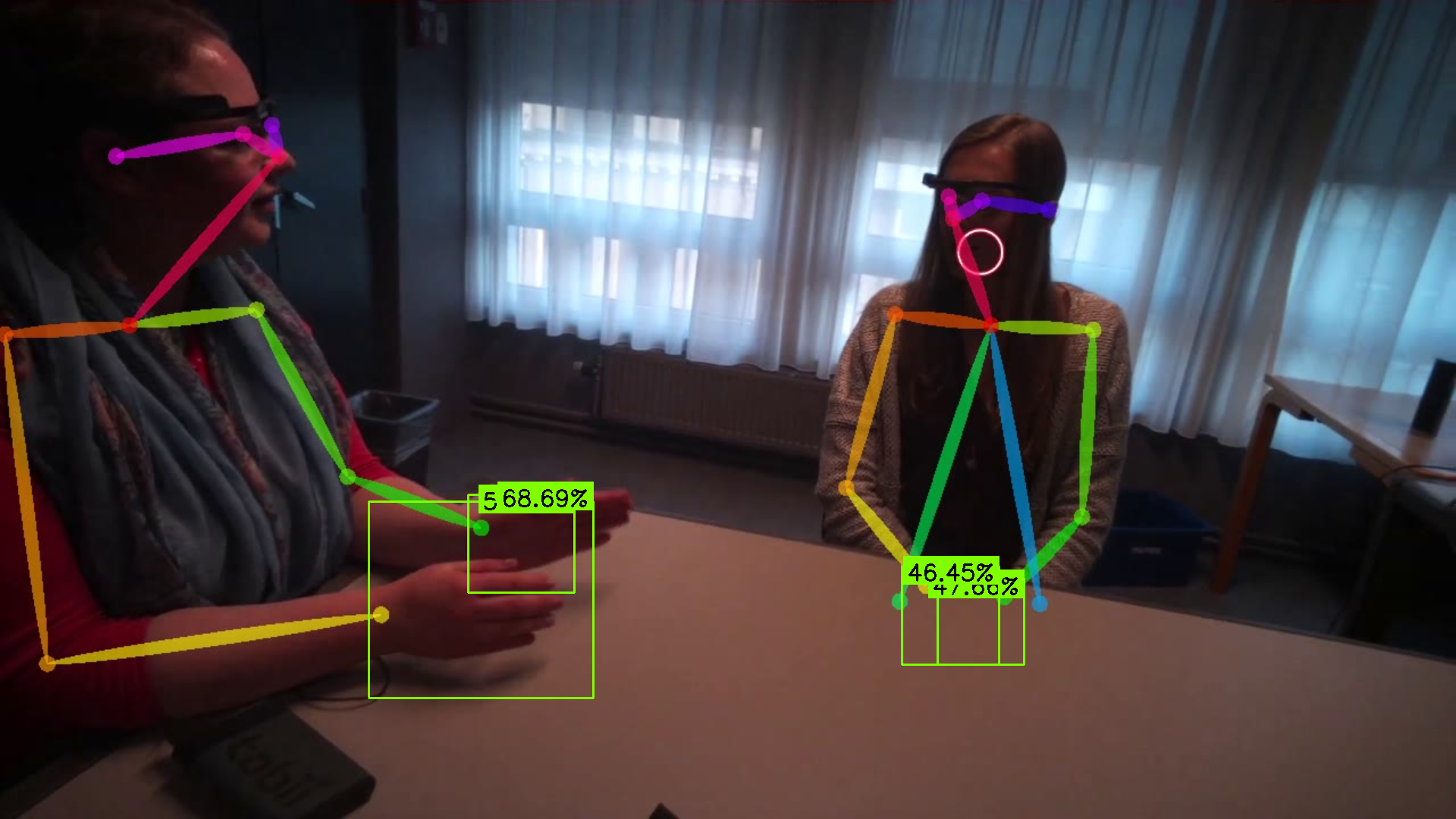}
  \caption{Hand estimated based on the elbow-wrist}
  \label{fig:hand_estimated}
\end{subfigure}
\caption{Three methods compared for hand detection}
\label{fig:hand_detections}
\end{figure}

Figure \ref{fig:hand_detections} illustrates the pose estimated hand, the hand pose detection and the YOLOv2 hand-head model detection.
Our initial intention was to combine the pose estimated hand detection with the hand pose detection, yet they seemed to contradict each other.
The estimated detection will be present even if the hands are occluded, as illustrated in figure \ref{fig:hand_estimated}, while figures \ref{fig:hand_pose} and \ref{fig:hand_yolo} show no detection or a very low detection confidence.
Both situations can be favourable depending on the aspect of the study and thus are complementary.

\section{Results}
\label{sec:results}

\subsection{Torso}

\begin{figure}
 \centering
   \begin{subfigure}{.49\textwidth}
 	\includegraphics[width=0.85\textwidth]{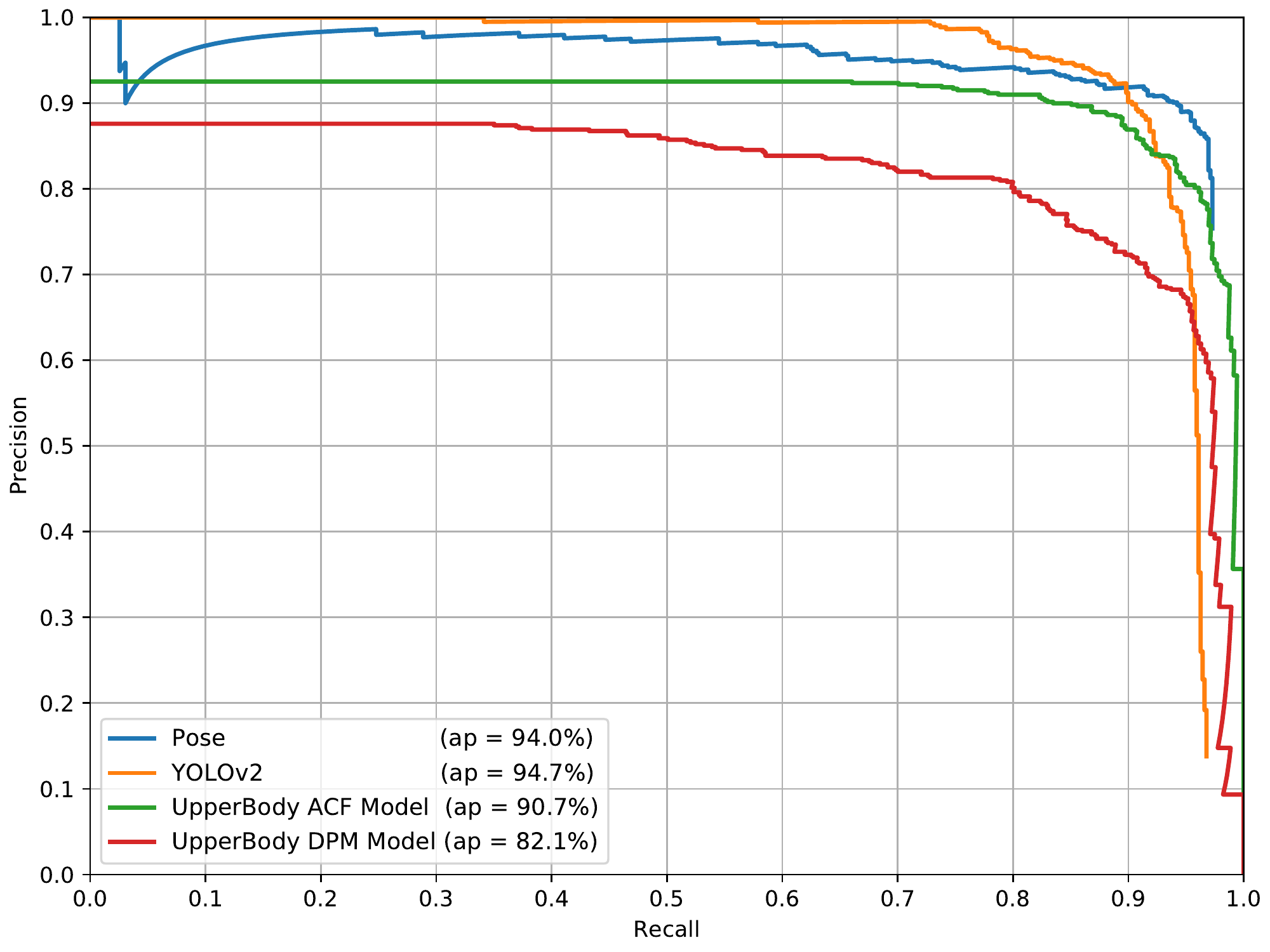}
 	\caption{Torso comparison on INRIA }
	\label{fig:torso_yolo_openpose_inria}
 	\end{subfigure}%
	\begin{subfigure}{.45\textwidth}
	\centering
	\includegraphics[width=0.85\textwidth]{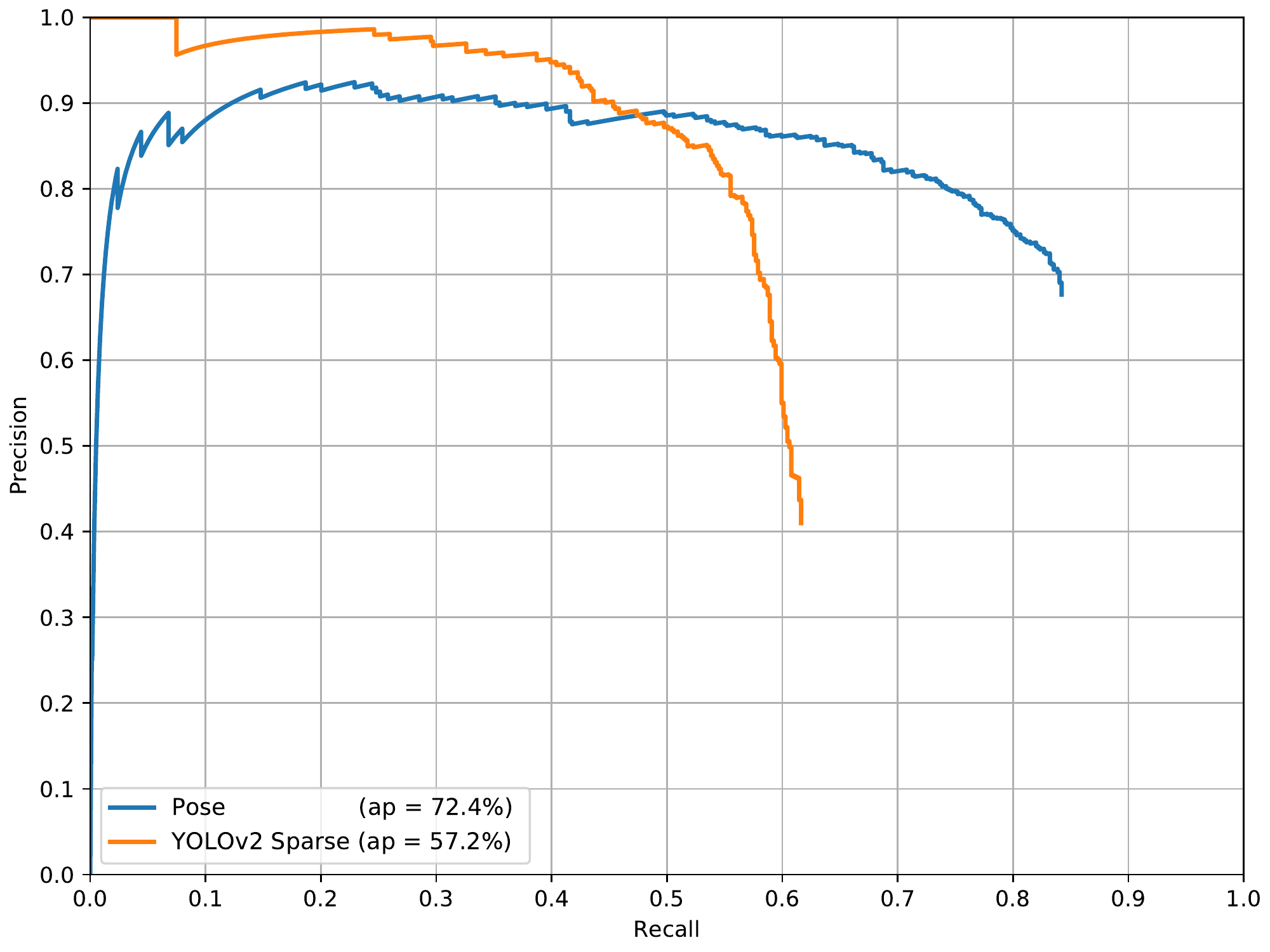}
	\caption{Head comparison on INRIA}
	\label{fig:head_yolo_openpose_inria}
	\end{subfigure}
 \vspace{-4mm}
\end{figure}

Our first technique concerns the torso detections compared to non deep learning techniques.
In the work of De Beugher et al. \cite{de2016computer}, two state-of-the-art upper-body models were tested on the INRIA person dataset \cite{dalal2005histograms} containing person annotations.
The dataset only contained full person detections, which was compensated for in \cite{de2016computer} by only taking into account the top 66\% of the person as upper body.
We evaluated our torso YOLOv2 model and pose-based torso detector on the same data.
Our results in the form of PR-curves are illustrated in figure \ref{fig:torso_yolo_openpose_inria} together with the UpperBody \gls{acf} \cite{yang2014aggregate} and UpperBody \gls{dpm} \cite{felzenszwalb2010cascade} models from \cite{de2016computer}.
These precision and recall curves were generated by varying a threshold over the detection confidence scores.
For the pose-based technique we have calculated the mean of all separate joint confidence scores.
Both the pose-based and YOLOv2 technique show an increased average precision opposed to the ACF and DPM model.

\subsection{Head}

Because only detecting the torso does not suffice as a basis for annotation, as mentioned above, we evaluated our head approach on the same INRIA dataset.
The INRIA dataset contained the head location and the person bounding box.
We used 66\% of the person annotation width as a reference for the head annotation size.
Figure \ref{fig:head_yolo_openpose_inria} illustrates the precision and recall of the YOLOv2 and pose-based model.
The recall drops faster compared to the results in figure \ref{fig:torso_yolo_openpose_inria}, which is to be expected since the head opposed to the full torso has a smaller area, increasing the detection challenge with a high confidence score.
Here the pose-based detection clearly outperforms the YOLOv2 model, mainly because the pose-based detections have the full yet hidden pose as decision support.
The YOLOv2 model has no such support and has to rely on the head only.

\subsection{Hands}
\begin{table}[tb]
\centering
\begin{tabular}{c|c|c|c|c|cccc|}
\cline{2-9}
                                & Mittal \cite{mittal2011hand} & Yang \cite{yang2011articulated}   & \multicolumn{2}{c|}{De Beugher \cite{de2015semi}}               & \multicolumn{4}{c|}{\textbf{Ours}}                                                                                                         \\ \cline{4-9}
                                &        &        &                             & incl. tracking & {Pose}                        & {Estimated}              & {Sparse}               & {Dense} \\ \hline
\multicolumn{1}{|c|}{D1}        & 85\%   & 24.2\% & \multicolumn{1}{c|}{83.4\%} & 88.2\%         & \multicolumn{1}{c|}{{98.4\%}} & \multicolumn{1}{c|}{{97.6\%}} & \multicolumn{1}{c|}{{92\%}}   & \textbf{99.4\%}       \\ \hline
\multicolumn{1}{|c|}{D2}        & 48.9\% & 46.5\% & \multicolumn{1}{c|}{52.9\%} & 65.3\%         & \multicolumn{1}{c|}{\textbf{91.1\%}} & \multicolumn{1}{c|}{{84.8\%}} & \multicolumn{1}{c|}{{48.2\%}} & {61.9\%}       \\ \hline
\multicolumn{1}{|c|}{5-Signers} & 77.6\% & n.a.   & \multicolumn{1}{c|}{81.1\%} & n.a.           & \multicolumn{1}{c|}{\textbf{97.6\%}} & \multicolumn{1}{c|}{{88.3\%}} & \multicolumn{1}{c|}{{84\%}}   & {92.2\%}         \\ \hline
\end{tabular}
\caption{F1-scores on the Insightout \cite{de2015semi} and 5-Signers \cite{buehler2008long} datasets}
\label{tbl:hands_compare}
\vspace{-7mm}
\end{table}

\begin{table}[tb]
\centering
\begin{tabular}{c|c|c|c|c|c|c|}
\cline{2-7}
                                            & Mittal \cite{mittal2011hand}    & Yang \cite{yang2011articulated}          & \multicolumn{1}{c|}{De Beugher \cite{de2015semi}} & \multicolumn{3}{c|}{\textbf{Ours}}                                \\ \cline{2-7}
                                            &                                 &         &                                & {Pose}  & {Estimated} & {YOLOv2 models} \\ \cline{1-7}
\multicolumn{1}{|c|}{Avg time/frame}        & 293.33 s                        & 113 s   & \multicolumn{1}{c|}{36.67 s}   & {0.5 s} & {0.125 s}        & \textbf{0.0099 s}      \\ \hline \cline{1-7}
\multicolumn{1}{|c|}{Avg fps} & 0.00341                         & 0.00885 & \multicolumn{1}{c|}{0.02750}   & { 2 } & {8}        & \textbf{100}      \\ \hline
\end{tabular}
\caption{Execution time comparison}
\label{tbl:hands_time}
\vspace{-7mm}
\end{table}

To evaluate the hand approaches we used the InsightOut dataset (D1 and D2) \cite{de2015semi} and the 5-Signers dataset \cite{buehler2008long}.
Table \ref{tbl:hands_compare} compares the F1-scores of the different approaches with our work.
Our results show that our proposed approaches all show an increase in F1-score opposed to existing models.
Only the YOLOv2 (\emph{Sparse, Dense}) models show a slight decrease in accuracy on the D2-dataset compared to the work of \cite{de2015semi}.

Besides the accuracy, we also compared the processing times of our techniques (Table \ref{tbl:hands_time}).
Previous techniques only used the CPU to process the data, whereas our approaches require a mid-end GPU (NVIDIA GTX 1080 Ti) capable of running the used algorithms.
We conclude that only the YOLOv2-based approach is able to run in real-time, although the proposed pose-based techniques are at least 70 times faster than the competitors.

\begin{figure*}
\centering
\begin{subfigure}{.3\textwidth}
  \centering
  \includegraphics[width=0.9\textwidth]{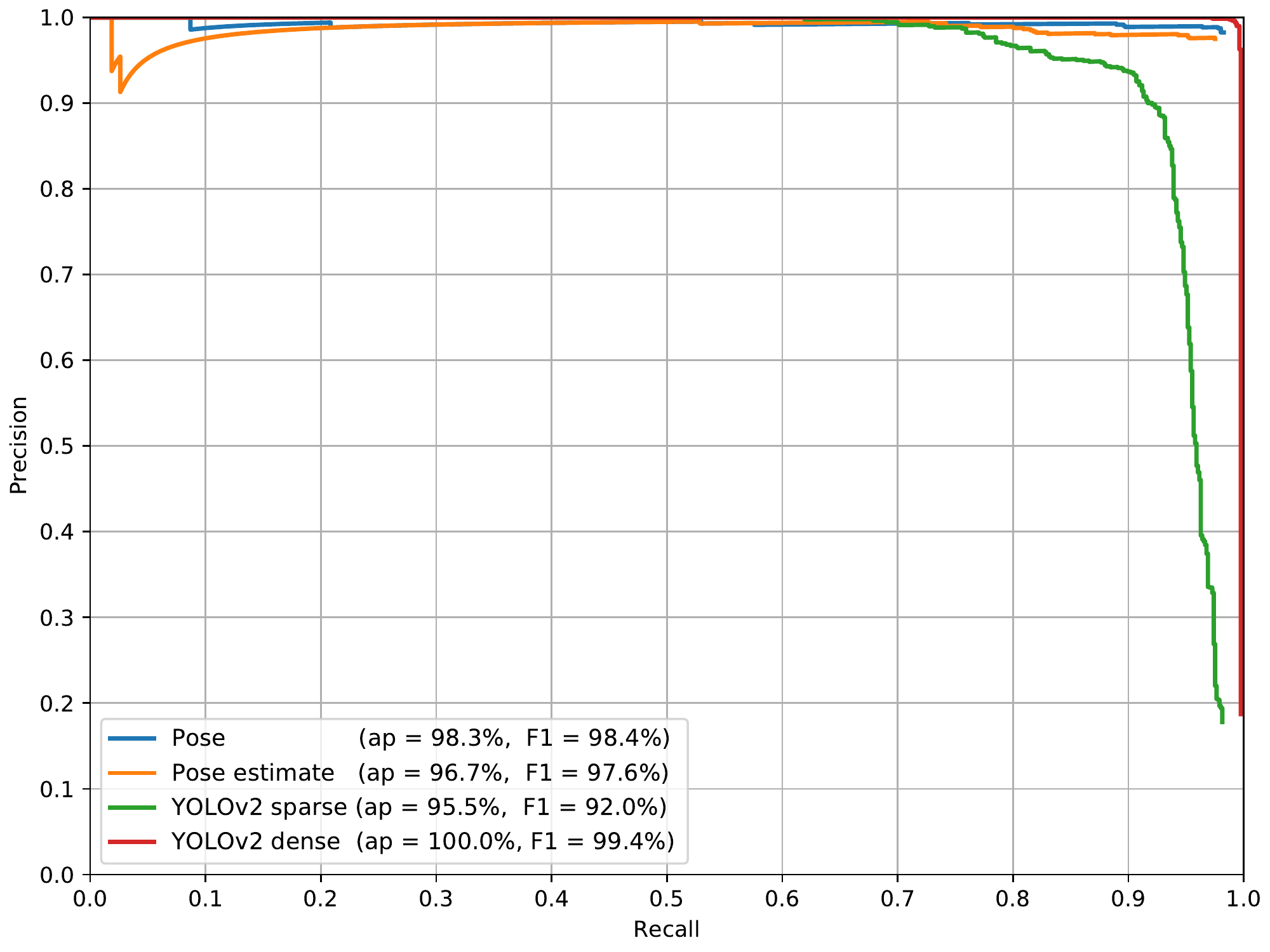}
  \caption{Hand results YOLOv2 and OpenPose on D1}
  \label{fig:hand_yolo_openpose_d1}
\end{subfigure}
\begin{subfigure}{.3\textwidth}
  \centering
  \includegraphics[width=0.9\textwidth]{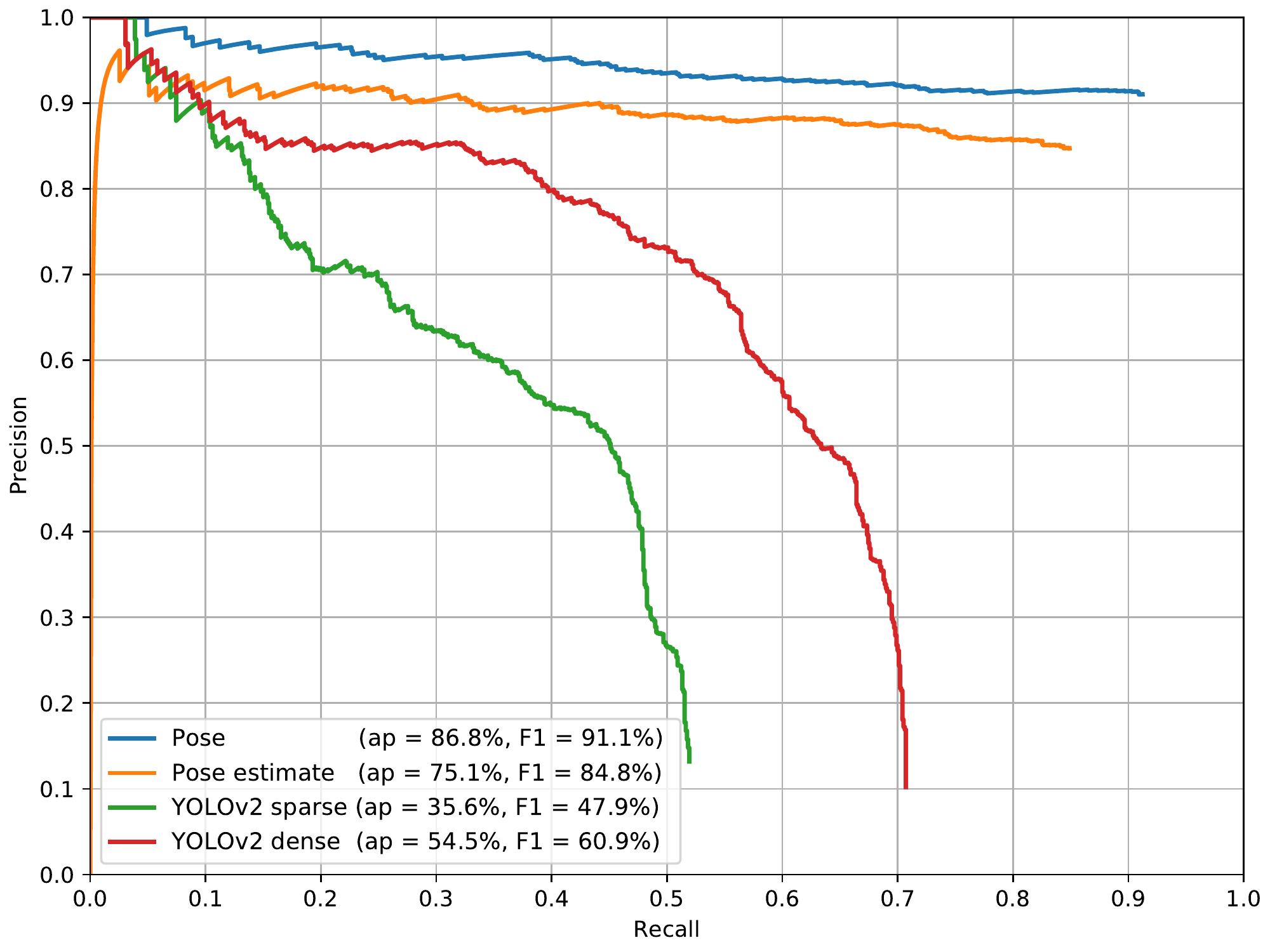}
  \caption{Hand results YOLOv2 and OpenPose on D2}
  \label{fig:hand_yolo_openpose_d2}
\end{subfigure}
\begin{subfigure}{.3\textwidth}
  \centering
  \includegraphics[width=0.9\textwidth]{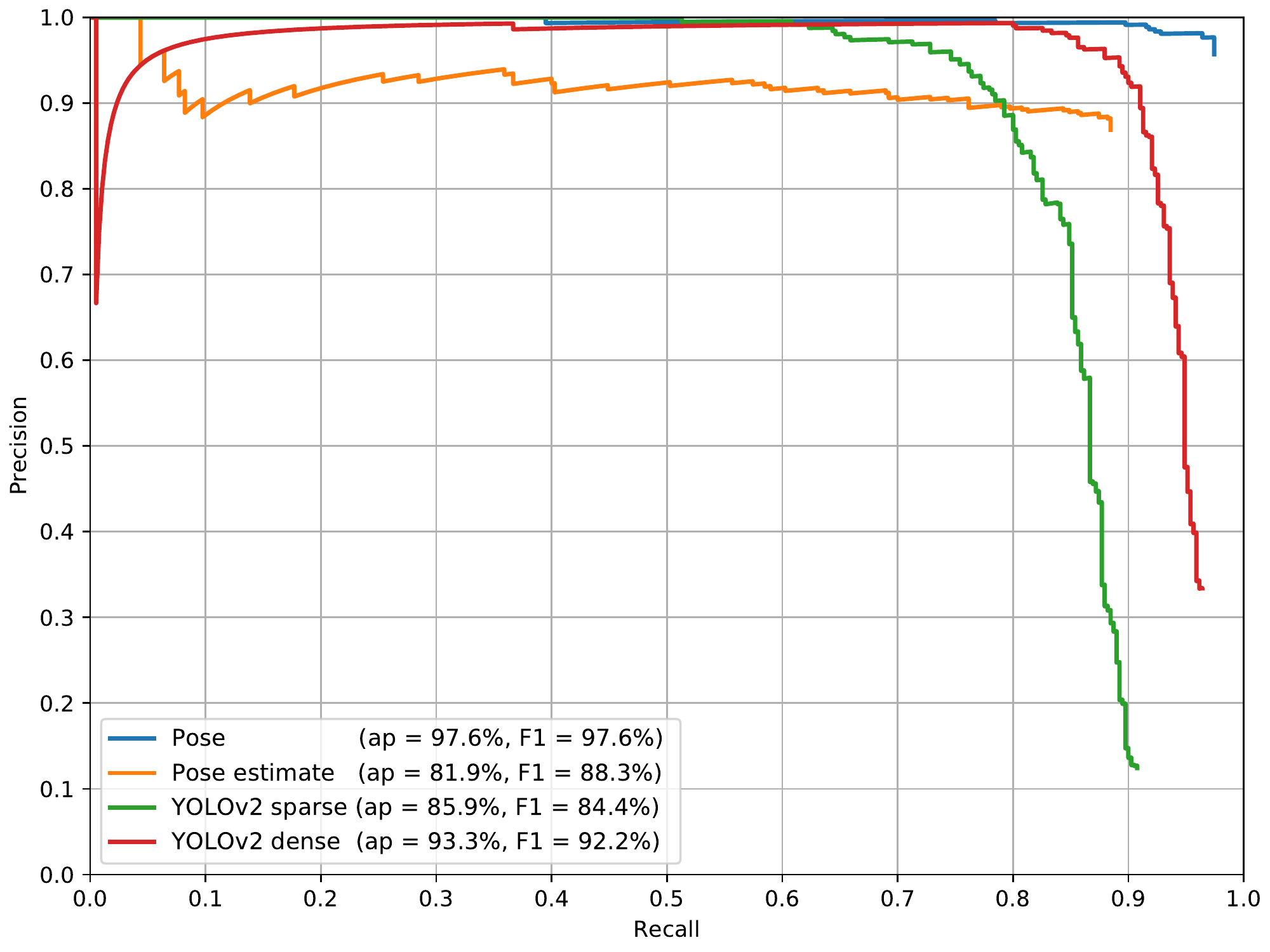}
  \caption{Hand results YOLOv2 and OpenPose on 5-Signers}
  \label{fig:hand_yolo_openpose_d3}
\end{subfigure}
\caption{Results on three hand datasets}
\label{fig:hand_compare}
\vspace{-8mm}
\end{figure*}

To compare these techniques against each other we plotted the PR-curve for each approach on each dataset, illustrated in figures \ref{fig:hand_yolo_openpose_d1}, \ref{fig:hand_yolo_openpose_d2} and \ref{fig:hand_yolo_openpose_d3}.
Comparing YOLOv2 against the pose-based techniques shows that both YOLOv2 models are less accurate on the D2 dataset.
We see that training a denser model on more hand data increases the mAP of the model compared to the sparse two-class model.

As expected we observe that the pose-based techniques produce good results.
The estimated hand location based on the elbow and wrist shows a decreased precision, which is expected due to the static direction on which we estimate the hand.
In reality the hand does not necessarily follow the arm movement explaining the performance drop compared to the hand pose estimator.

\subsection{Automated Annotations}

The main goal of our work, providing automatic annotations during human-human interactions, will produce labels on the recordings, depending on overlap between the detections and the gaze.
When the gaze falls within the boundaries of a detection, the detection label will be the coded for that frame.
To evaluate the head gaze labels, we manually annotated 2500 frames of a face-to-face spontaneous conversation between three people wearing mobile eye-trackers.
On this segment we used both the OpenPose head detector and YOLOv2 sparse head model to generate automatic labels for each frame.
In case of manual annotation of recordings it may occur that opinions differ between annotators.
To overcome this source of discussion the annotations are commonly compared by different measures to obtain a score of resemblance (referred to as an inter-coder agreement test or ICA-test)
In order to compare our automated labels with the ground truth we use the same tests to obtain scores evaluating our techniques.
These results are visible in table \ref{tbl:head_ica_levels}.
The annotated video of this sequence, using the Pose based head and hand detection can be viewed on https://youtu.be/eEVXIfY99O0.

\begin{table}[]
\centering
\begin{tabular}{l|l|l|}
\cline{2-3}
                                          & OpenPose Level & YOLOv2 Level \\ \hline
\multicolumn{1}{|l|}{Agreement \cite{neuendorf2016content}}             & 91.1\%         & \textbf{92.6\%}     \\ \hline
\multicolumn{1}{|l|}{Scott's Pi \cite{scott1955reliability}}            & 82.2\%         & \textbf{85.3\%}     \\ \hline
\multicolumn{1}{|l|}{Cohen's Kappa \cite{cohen1960coefficient}}         & 82.3\%         & \textbf{85.3\%}     \\ \hline
\multicolumn{1}{|l|}{Krippendorf's Alpha \cite{krippendorff2012content}} & 82.2\%         & \textbf{85.3\%}     \\ \hline
\end{tabular}
\caption{Reliability levels of the automated head annotation }
\label{tbl:head_ica_levels}
\vspace{-5mm}
\end{table}

\section{Conclusion}
\label{sec:conclusion}

This paper focussed on comparing the current state-of-the-art techniques on automated mobile eye-tracking analysis.
This involves detecting the hands and heads appearing in the scene images generated by the eye-tracker, which may be obvious foci of attention during face-to-face human interactions.
The output of this automatic detection step provides a solid basis for further annotation of relevant non-verbal behaviour, including hand gestures, head movements and so on.
We compared two main techniques, a pose estimator and the YOLOv2 detector against more traditionally used techniques showing an overall higher accuracy.
Despite the fact that these deep learning techniques require a mid-end GPU, they easily achieve faster than real-time performance.
By providing multiple techniques and allowing adjustable bounding boxes margins, the gaze annotations are customizable according to the requirements of the specific study at hand without any need for manual intervention.

\bibliographystyle{splncs}
\bibliography{bibliography.bib}

\end{document}